\documentclass[10pt,twocolumn,letterpaper]{article}
\usepackage{preprint}
\definecolor{linkblue}{rgb}{0.21,0.49,0.74}
\usepackage[breaklinks,colorlinks,allcolors=linkblue]{hyperref}
\hypersetup{
    pdftitle={A Generative AI Agent to Emulate Interactive Human-driven Vehicles for Automated Driving Testing in Urban Areas},
    pdfauthor={Jia Hu, Changjian Yu, Panqu Wang, Jaehyun So, Yang Chang, Xuerun Yan, Jintao Lai}
}

\title{A Generative AI Agent to Emulate Interactive Human-driven Vehicles for Automated Driving Testing in Urban Areas}

\author{
Jia Hu$^{1}$ \quad Changjian Yu$^{1}$ \quad Panqu Wang$^{2}$ \quad Jaehyun So$^{3}$\\
Yang Chang$^{1}$ \quad Xuerun Yan$^{1}$ \quad Jintao Lai$^{1,*}$\\[3pt]
$^{1}$Tongji University \quad $^{2}$ZERON \quad $^{3}$Ajou University\\[4pt]
{\small $^{*}$Corresponding author: \texttt{Jintao.Lai@tongji.edu.cn}}
}
\date{}

\begin{document}
\maketitle

\begin{abstract}
Realistic modeling of human-driven vehicle (HDV) interactions is essential for automated driving safety testing in urban areas. Existing methods struggle to reproduce realistic safety-critical interactions involving high collision risk for the following reasons: they often assume full observability, overlooking interactions arising from occlusion. Driving styles are mainly characterized by vehicle maneuvers, limiting the diversity of safety-critical interaction behaviors. Moreover, learning signals from safety-critical interactions can be diluted by non-critical driving behaviors. To address these issues, we propose a generative AI agent to emulate interactive HDVs. The agent incorporates field-of-view modeling to account for driver visibility under occlusion and characterizes the interaction process using five metrics spanning its initial, intermediate, and final stages. It automatically learns diverse interaction styles through a dynamic mixture-of-experts network. Through residual reinforcement learning, the agent further refines generated trajectories to match the real-world distributions of safety-critical interaction characteristics. Experiments on the INTERACTION dataset show that our method reduces Jensen--Shannon divergence (JSD) by 39.9\%--87.1\% across five interaction characteristics compared with a state-of-the-art baseline. Our method achieves greater improvements of up to 87.1\% in final-stage interaction metrics, suggesting that it can reduce error accumulation in long-horizon trajectory generation. Ablation studies confirm the effectiveness of each proposed component, particularly highlighting the importance of residual refinement. Generalization experiments show that our method approaches the performance of full training on unseen intersections with only one epoch of fine-tuning.
\end{abstract}
\hypertarget{introduction}{%
\section{Introduction}\label{introduction}}

\begingroup
\setlength{\parskip}{0pt}

As automated driving systems move toward deployment in urban environments, validating the safety of autonomous vehicles (AVs) in complex traffic becomes increasingly important \citep{Feng2021NADE,Hu2023UrbanCACC}.
Compared with highways, urban areas involve more traffic conflicts from multiple directions, especially at intersections \citep{Zhan2019INTERACTION}.
Buildings and surrounding vehicles can also create blind spots, making it difficult for AVs to detect and respond to other vehicles in a timely manner \citep{Shao_2023_CVPR,Wang2023OcclusionPlanning}.
These challenges increase the risk of traffic accidents.
Therefore, it is essential to validate whether AVs can safely handle these complex traffic conditions in urban areas.

To test and validate AV safety, replicating the interactions between AVs and human-driven vehicles (HDVs) is critical.
In real traffic, human drivers react to an AV's actions, and the AV adjusts its behavior in response \citep{Bergamini2021SimNet,Suo_2021_CVPR}.
Safety tests must capture this interaction process to reflect the challenges of driving among human drivers.
Without realistic interactions, test results may give a misleading picture of AV safety \citep{Feng2021NADE}.
Therefore, replicating these interactions is essential for assessing how safely AVs can operate on real roads.

To reproduce these interactions, adversarial driving behavior modeling for HDVs in simulation environments is necessary.
Some interactions involve high collision risks but occur rarely \citep{Liu2024Rarity,Feng2023DenseRL}.
Such safety-critical interactions are difficult to reproduce safely through on-road testing \citep{Liu2024Rarity,Feng2023DenseRL}.
Simulation allows them to be generated safely and repeatedly \citep{Rempe_2022_CVPR}.
Within simulation environments, adversarial driving behavior models for HDVs generate challenging behaviors to test how AVs respond to risks \citep{Feng2021NADE,Feng2023DenseRL,Wang2021AdvSim}.

For adversarial driving behavior modeling, early studies mainly used heuristic-based models.
The behaviors are generated by predefined rules, leading to rigid and unrealistic trajectories.
Typical examples include car-following and lane-changing models \citep{Treiber2000IDM,Kesting2007MOBIL}.
Such models struggle to capture complex interactions, including yielding and negotiating right-of-way at urban intersections \citep{Suo_2021_CVPR}.
As a result, the generated trajectories often lack the flexibility and realism of human driving.

To improve realism, imitation-based methods learn HDV behaviors from real-world data, but their training is often dominated by common driving patterns, resulting in limited behavior diversity \citep{wang2025hadgen,Lian_2026_CVPR}.
Common behaviors provide more training examples and have greater influence on the learned model.
Less frequent interaction patterns may be poorly represented, even when average trajectory accuracy is high \citep{Liu2024Rarity}.
This reduces the diversity of driving behaviors.

More recently, generative methods construct an agent that produces diverse and plausible HDV behaviors to enhance diversity \citep{Wu2024SMART,Lin2025CCDiff,Chang2024SafeSim}.
However, the realism of safety-critical interaction remains limited due to three key challenges:

\begin{enumerate}[nosep,label=\arabic*)]
    \item
    \textbf{They assume full observability without occlusion, overlooking risky interactions that arise due to blind spots.}
    At urban intersections, buildings and surrounding vehicles may obstruct an HDV's view of approaching traffic.
    With access to all surrounding vehicles, the model may generate reactions to vehicles that the HDV cannot actually see, resulting in unrealistic interaction behaviors.

    \item
    \textbf{They mainly characterize driving styles based on vehicle driving maneuvers rather than interactions, which makes it difficult to distinguish different interaction patterns.} 
    This limits the diversity of safety-critical interactions. For example, drivers performing the same left-turn maneuver may differ in their yielding decisions and accepted gaps \citep{wang2023safety}.
    Such differences reflect distinct interaction behaviors that cannot be represented by driving maneuvers \citep{lan2026controllable}.

    \item
    \textbf{The learning signal from safety-critical interactions can be diluted by that from non-critical driving behaviors, leading to a mismatch between generated and real-world behavioral distributions.}
    Safety-critical interactions often occupy a small part of a trajectory \citep{Feng2023DenseRL}.
    A model may therefore achieve high overall trajectory accuracy without matching the distributions of critical interaction characteristics \citep{Zhang_2025_CVPR,Ahmadi_2026_CVPR}.
    Such mismatches can misrepresent the risks AVs face during testing and make safety evaluations less reliable \citep{Feng2021NADE}.
\end{enumerate}

Thus, this study proposes a generative AI agent to emulate the interactive behaviors of HDVs for automated driving testing in urban areas.
The proposed method has the following features:

\begin{itemize}[nosep,topsep=3pt]
    \item 
    \textbf{Reproduce occlusion-constrained interaction under partial observability.}
    \item 
    \textbf{Improve diversity of interactive behaviors.}
    \item 
    \textbf{Match the real-world distributions of safety-critical interaction characteristics.}
\end{itemize}

To achieve these features, the proposed method has three key technical highlights:

\begin{itemize}[nosep,topsep=3pt]
    \item
    \textbf{Introduce vehicle field-of-view-based visibility to learn occlusion-constrained interaction:}
    A field-of-view (FOV) mask is constructed to identify surrounding vehicles observable to the target HDV.
    This mask filters out occluded vehicles during interaction encoding, allowing the model to generate driving behaviors based on the visible traffic context.

    \item
    \textbf{Design a dynamic expert-activation mixture-of-experts (MoE) network for learning interaction styles:}
    Different trajectory experts are dynamically activated based on the interaction context. These experts learn distinct interaction patterns from real-world driving data, enabling the generation of diverse HDV behaviors under similar traffic conditions.

    \item
    \textbf{Introduce sparse rewards to learn the diluted signals of safety-critical interactions:}
    Sparse rewards are designed based on critical interaction characteristics to provide targeted learning signals.
    Group Relative Policy Optimization (GRPO) \citep{Shao2024GRPO} is employed to train a residual policy that refines generated trajectories toward the real-world distributions of these characteristics.
\end{itemize}

\par
\endgroup
\section{Related Work}
\label{related-work}

\subsection{Traffic Agent Simulation}
\label{rw-traffic-simulation}

Traffic agent simulation aims to reproduce the behaviors of surrounding road users for autonomous driving testing \citep{Suo_2021_CVPR,Bergamini2021SimNet}.
Traffic simulation also supports the evaluation of cooperative cruise control and truck platooning in mixed traffic \citep{Lai2020GenericCACC,Hu2023UrbanCACC,Hu2024TruckPlatooning}.
Early approaches model car-following and lane-changing behaviors through predefined rules, but struggle to capture complex human driving interactions \citep{Treiber2000IDM,Kesting2007MOBIL,Suo_2021_CVPR}.
Learning-based methods address this limitation by learning traffic behaviors from real-world data \citep{Xu2023BITS}.
TrafficSim uses a latent-variable model to generate joint multi-agent behaviors, while SimNet learns reactive vehicle behaviors from recorded observations \citep{Suo_2021_CVPR,Bergamini2021SimNet}.
Waymax supports closed-loop evaluation from recorded scenes, and MixSim makes such scenes reactive and controllable \citep{Gulino2023Waymax,Suo2023MixSim}.
RTR further combines imitation and reinforcement learning to improve traffic compliance in closed-loop simulation \citep{Zhang2023RTR}.
Imitation and reinforcement learning have also been combined for urban trajectory planning \citep{Zeng2026RITP}.
More recently, generative methods have enabled more flexible behavior generation and control \citep{Zhong2023CTG,Lin2025CCDiff}.
SceneGen and TrafficGen learn to generate traffic scenes from driving data \citep{Tan2021SceneGen,Feng2023TrafficGen}.
Other generators target dense agent interactions and city-scale traffic \citep{yang2025hid2,Tan2025SceneDiffuserPP}.
ProSim uses user prompts to control agent behaviors in closed-loop traffic simulation \citep{Tan2025ProSim}.
SAFE-SIM further employs diffusion models to generate controllable adversarial behaviors for safety-critical testing \citep{Chang2024SafeSim}.
Together, these studies have advanced the development of reactive and controllable traffic agents, providing a foundation for realistic AV--HDV interaction simulation.

\subsection{Realistic Driving Behavior Modeling}
\label{rw-realistic-behavior}

Realistic driving behavior modeling aims to capture both the diversity of human driving behaviors and the statistical patterns of their interactions \citep{lan2026controllable}.
Interactions between vehicles also inform trajectory prediction \citep{Meng2026GraphInteraction,Yan2026TrajectoryReview,Ngiam2022SceneTransformer,Jiang2023MotionDiffuser} and lane-change planning \citep{Zhang2026Stackelberg,Wang2022MakeSpace}.
TrafficBots introduces a latent personality to represent agent-specific driving behaviors, while HAD-Gen identifies driving styles from real-world trajectories and learns style-specific driving policies \citep{Zhang2023TrafficBots,wang2025hadgen}.
Beyond behavioral diversity, recent studies have emphasized the statistical realism of generated traffic behaviors \citep{Zhang_2025_CVPR}.
NeuralNDE models normal and safety-critical driving events and calibrates their occurrence frequencies to match real-world statistics \citep{Yan2023NeuralNDE}.
WOSAC evaluates traffic agent realism through distributions of motion, interaction, and map-related features, while RLFTSim employs reinforcement learning fine-tuning to improve distributional realism \citep{Montali2023WOSAC,Ahmadi_2026_CVPR}.
SMART \citep{Wu2024SMART} discretizes vehicle motion into tokens and generates multi-agent trajectories through autoregressive next-token prediction.
The original study reported state-of-the-art performance on most metrics of the Waymo Open Sim Agents Challenge.
However, realistic safety-critical interactions under occlusion remain challenging. Our work addresses this gap through realistic HDV interaction generation.

\begin{figure*}[t]
\centering
\includegraphics[width=\linewidth]{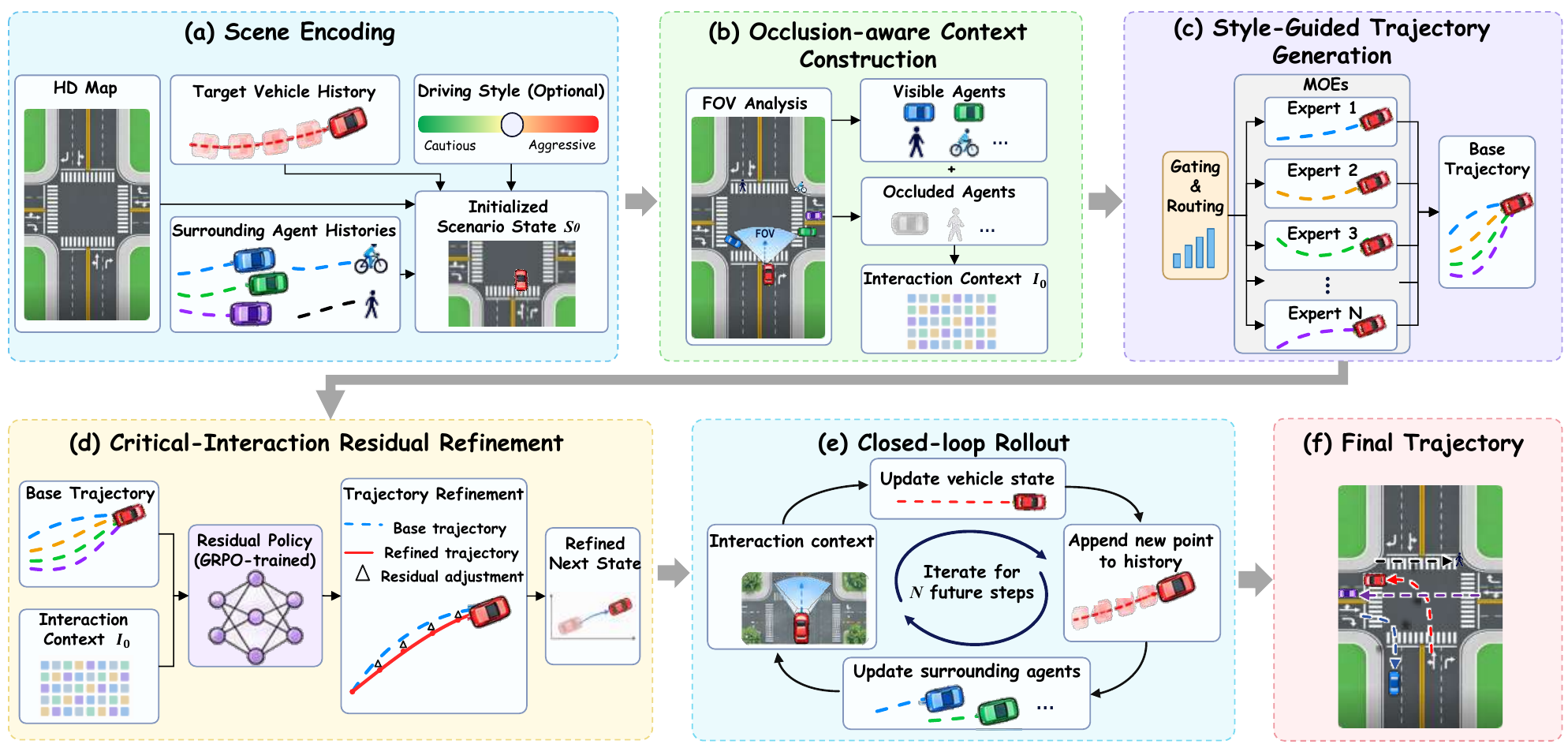}
\caption{Overview of the generative agent. Occlusion-aware Context Construction encodes visible traffic for Style-Guided Trajectory Generation. Critical-Interaction Residual Refinement uses a GRPO-trained residual policy to correct the base trajectory. The refined state updates the history for the next rollout step.}
\label{fig:technical_route}
\end{figure*}
\section{Method}
\label{method}

In this section, we introduce four key components of the proposed method: problem formulation (Sec.~3.1), occlusion-aware context construction (Sec.~3.2), style-guided trajectory generation (Sec.~3.3), and critical-interaction residual refinement (Sec.~3.4).

\subsection{Problem Formulation}
\label{problem-formulation}

Let $s_t=(p_t,v_t)$ denote the position and velocity of an HDV. Its history is $H_t=\{s_{t-H+1},\ldots,s_t\}$, and $\mathcal N$ denotes the states of the surrounding agents. Given a vectorized road map $M$, an interaction-style code $z$, and a prescribed goal position $g$, the agent generates a future trajectory
\begin{equation}
\hat\tau\sim p_\Theta(\cdot\mid H_0,\mathcal N,M,z,g),
\label{eq:trajectory-generation}
\end{equation}
\begin{equation}
\hat\tau=(\hat s_1,\ldots,\hat s_T),
\label{eq:trajectory-definition}
\end{equation}
where $H$ and $T$ denote the history and generation horizons. The model constructs a visibility-dependent context, generates an expert-conditioned base motion, and refines it before updating the target state.

\subsection{Occlusion-aware Context Construction}
\label{occlusion-aware-interaction-context-construction}

The occlusion-sensitive context construction encodes the traffic information observable to the target HDV. We compute its field of view (FOV) using a target-centered occupancy representation and shadowcasting. Vehicle occupancy blocks the line of sight to vehicles behind them, producing a binary visibility indicator $\nu_t^j$ for each surrounding vehicle $j$. Combined with the vehicle-existence indicator $E_t^j$, this produces the attention mask $m_t^j=E_t^j\nu_t^j$.

The target history, surrounding histories, and road-map polylines are encoded separately. The target-history embedding initializes a query that includes surrounding vehicles, the terminal-state condition, and map features. Vehicle attention incorporates the FOV mask:
\begin{equation}
\operatorname{Attn}(Q,K,V;m_t)
=\operatorname{softmax}\!\left(
\frac{QK^\top}{\sqrt d}+\log m_t
\right)V,
\label{eq:fov-attention}
\end{equation}
where $Q$, $K$, and $V$ are query, key and value features, and $d$ is the key dimension. With $\log 0=-\infty$, occluded vehicles receive zero attention weight. If no surrounding vehicle is visible, the vehicle-attention output is set to zero. Residual attention blocks fuse visible traffic, target history, and road geometry into the context $h_t$. This makes the generated response depend on the target driver's observable surroundings.

\subsection{Style-Guided Trajectory Generation}
\label{interaction-style-guided-trajectory-generation}

Style-guided trajectory generation employs imitation learning to capture diverse HDV interaction behaviors from real-world demonstrations. At each time step, the
visibility-dependent interaction context $h_t$ from
Sec.~\ref{occlusion-aware-interaction-context-construction}
is used for both expert routing and displacement prediction.
Following DynMoE \citep{Guo2025DynMoE}, the model adapts the
number of experts activated for each context and the total
number of experts during training.

Let $K$ be the current number of experts. For $k=1,\ldots,K$,
let $D_k$ denote the $k$-th trajectory expert, $w_k$ its routing
vector, and $\vartheta_k$ its learned activation threshold.
The router measures the similarity between $h_t$ and $w_k$:
\begin{equation}
c_k(h_t) =
\frac{h_t^\top w_k}{\|h_t\|\,\|w_k\|+\epsilon_g},
\end{equation}
where $\epsilon_g>0$ provides numerical stability. Each expert
is activated when its score exceeds its own threshold:
\begin{equation}
a_k(h_t) =
\mathbf{1}\!\left[
\sigma(c_k(h_t)) > \sigma(\vartheta_k)
\right],
\end{equation}
where $\sigma$ is the sigmoid function. The set of active experts is
\begin{equation}
\mathcal{A}(h_t) = \{k : a_k(h_t)=1\}.
\end{equation}
The independent gates allow different contexts to activate different numbers of experts. A straight-through estimator provides gradients for the binary gates during training.

During training, we set $\mathcal{S}_t=\mathcal{A}(h_t)$.
Each selected expert predicts a displacement from $h_t$.
Their predictions are averaged to obtain the base displacement:
\begin{equation}
b_t =
\frac{\displaystyle\sum_{k\in\mathcal{S}_t}D_k(h_t)}
{\max(1,|\mathcal{S}_t|)}.
\label{eq:dynamic-expert-generation}
\end{equation}

The expert pool is adjusted using routing records collected during training. Contexts that activate no existing expert are collected over a training window. Let $r$ be the sum of their
context vectors. When $r\ne 0$, a new expert is added with routing vector $w_{\mathrm{new}}=r/\|r\|$ and initial threshold
$\vartheta_{\mathrm{new}}=0$. Experts that receive no assignments during the window are removed.

The generator learns to match the recorded displacement between
consecutive time steps:
\begin{equation}
\begin{aligned}
\mathcal{L}_{\mathrm{IL}}
={}&\mathbb{E}_{\tau\sim\mathcal{D}}\!\left[
\frac{1}{T}\sum_{t=0}^{T-1}
\ell_{\mathrm{disp}}(b_t,p_{t+1}-p_t)
\right]\\
&+\lambda_{\mathrm{reg}}\mathcal{L}_{\mathrm{reg}},
\end{aligned}
\label{eq:imitation-objective}
\end{equation}
where $\mathcal{D}$ is the set of recorded demonstrations,
$p_t$ is the recorded position at time $t$, and
$\ell_{\mathrm{disp}}$ measures displacement error.
The regularization term combines trajectory smoothness, endpoint
consistency, spatial penalties, and routing regularization.
The displacement loss trains the selected experts to imitate recorded motion, while routing organizes the demonstrations across experts.

Through imitation learning, different sets of experts learn to reproduce different interaction patterns. We use $z$ to index a learned interaction style and denote its associated expert set by $\mathcal{S}_{z}$. During generation, the selected style fixes the expert set throughout the rollout. At each step, the experts in $\mathcal{S}_{z}$ predict displacements from the current context $h_t$. Their mean gives the base displacement $b_t$, which is passed to the residual refinement module.

\subsection{Critical-Interaction Residual Refinement}
\label{critical-interaction-aware-trajectory-refinement}

Critical-interaction residual refinement learns corrections to the base motion through reinforcement learning with critical-interaction rewards. The base generator remains fixed, and a residual policy learns bounded displacement corrections. Let $x_t$ summarize the target state and its generated history. The policy samples
\begin{equation}
u_t\sim\pi_\psi(\cdot\mid x_t)
=\mathcal N\!\left(\mu_\psi(x_t),
\operatorname{diag}(\sigma_\psi^2(x_t))\right),
\label{eq:residual-sampling}
\end{equation}
\begin{equation}
\delta_t=\operatorname{clip}
(u_t,-\delta_{\max},\delta_{\max}),
\label{eq:residual-correction}
\end{equation}
\begin{equation}
a_t=\operatorname{clip}
(b_t+\delta_t,-a_{\max},a_{\max}),
\label{eq:refined-displacement}
\end{equation}
where $a_t$ is the refined displacement. The bounds $\delta_{\max}$ and $a_{\max}$ control the correction and final displacement magnitudes. The base motion carries the visibility-dependent scene information, while the residual policy adjusts its continuation.

Reward construction focuses on the interaction characteristics defined in Sec.~\ref{experiment-setup}. Let $\xi_e(\hat\tau)$ denote the value of interaction characteristic $e$, and $f_e$ its empirical histogram frequency in the training data. The event reward is
\begin{equation}
R_{\mathrm{event}}(\hat\tau)
=\sum_{e\in\mathcal E_{\mathrm{int}}}
w_e f_e\!\left[\operatorname{bin}(\xi_e(\hat\tau))\right].
\label{eq:interaction-event-reward}
\end{equation}
Here, $\mathcal E_{\mathrm{int}}$ denotes the set of interaction characteristics used for reward computation. This targeted feedback is combined with vehicle proximity and road geometry penalties:

\begin{equation}
R(\hat\tau)=R_{\mathrm{event}}(\hat\tau)
-\lambda_C P_C(\hat\tau)
-\lambda_B P_B(\hat\tau).
\label{eq:refinement-reward}
\end{equation}
Here, $P_C$ penalizes close vehicle proximity, while $P_B$ penalizes proximity to sampled map points. These penalties complement the event reward by encouraging safe and spatially valid trajectories.

We adapt Group Relative Policy Optimization (GRPO) \citep{Shao2024GRPO} to train the residual policy. For the same initial scene, the sampling policy $\pi_{\psi_{\mathrm{old}}}$ generates a group of $G$ candidate continuations. Their rewards $R_i=R(\hat\tau_i)$ provide group-relative advantages
\begin{equation}
A_i=\frac{R_i-\mu_R}{\sigma_R+\epsilon_s},
\label{eq:group-advantage}
\end{equation}
where $\mu_R$ and $\sigma_R$ are the group reward mean and standard deviation, and $\epsilon_s>0$ provides numerical stability. Let $U_i=(u_{i,0},\ldots,u_{i,T-1})$ be the sampled residual-action sequence for candidate $i$. Its states $x_{i,t}$ are obtained from its own rollout. We use a trajectory-level likelihood ratio
\begin{equation}
\rho_i(\psi)=
\prod_{t=0}^{T-1}
\frac{\pi_\psi(u_{i,t}\mid x_{i,t})}
{\pi_{\psi_{\mathrm{old}}}(u_{i,t}\mid x_{i,t})}
\label{eq:trajectory-ratio}
\end{equation}
in the clipped objective
\begin{equation}
\begin{aligned}
\mathcal J_{\mathrm{RL}}(\psi)
=\mathbb E\Bigg[
&\frac{1}{G}\sum_{i=1}^{G}
\min\!\left(\rho_i A_i,\bar\rho_i A_i\right)\\
&+\frac{\lambda_H}{GT}\sum_{i,t}
\mathcal H\!\left(\pi_\psi(\cdot\mid x_{i,t})\right)
\Bigg],
\end{aligned}
\label{eq:grpo-refinement}
\end{equation}
\begin{equation}
\bar\rho_i=\operatorname{clip}
(\rho_i,1-\epsilon_c,1+\epsilon_c).
\label{eq:policy-ratio-clipping}
\end{equation}
The expectation is over training scenes and candidate groups sampled from $\pi_{\psi_{\mathrm{old}}}$. Here $\epsilon_c$ controls policy-ratio clipping, and $\mathcal H$ is policy entropy with coefficient $\lambda_H$. Group comparison assigns higher relative advantages to continuations with stronger interaction rewards under the same scene conditions.

\textbf{Closed-loop Rollout.}
The refined displacement updates position and velocity as $p_{t+1}=p_t+a_t$ and $v_{t+1}=a_t/\Delta t$, where $\Delta t$ is the integration interval. The new state is appended to the target history. The agent then recomputes visibility and context from the updated scene and generates the next displacement. Repeating this process produces the future trajectory over horizon $T$.

\begingroup
\clubpenalty=10000
\widowpenalty=10000

\hypertarget{experiment}{%
\section{Experiments}\label{experiment}}

\hypertarget{experiment-setup}{%
\subsection{Experiment setup}\label{experiment-setup}}

\textbf{Datasets.}
We use vehicle interaction data from two distinct intersections in the INTERACTION dataset \citep{Zhan2019INTERACTION}. DR\_USA\_Intersection\_GL (GL) is used for the main evaluation, while DR\_USA\_Intersection\_MA (MA) serves as the source for generalization experiments. 

\textbf{Comparison configurations.}
The baseline is SMART \citep{Wu2024SMART}, labeled \emph{SOTA} in figures and tables. It uses placeholder map inputs and motion-token decoding. \emph{Ours} denotes the complete agent. \emph{With/Without FOV} and \emph{With/Without style module} identify component ablations. \emph{IL} denotes imitation learning alone; \emph{PPO} and \emph{GRPO (Ours)} add residual-policy refinement.

\textbf{Metrics.}
We evaluate interaction realism using five characteristics: initial time-to-collision ($\mathrm{TTC}_0$), initial radial relative speed ($\mathrm{RS}_0$), adjustment duration ($T_{\mathrm{adj}}$), final time-to-collision ($\mathrm{TTC}_1$), and final radial relative speed ($\mathrm{RS}_1$). $\mathrm{TTC}_0$ and $\mathrm{TTC}_1$ measure the estimated time to collision at the beginning and end of speed adjustment, respectively. $\mathrm{RS}_0$ and $\mathrm{RS}_1$ describe the relative motion between interacting vehicles at these two moments, with positive values indicating approach and negative values indicating separation. $T_{\mathrm{adj}}$ measures the normalized duration of speed adjustment. These interaction characteristics are also used to construct the critical-interaction rewards in Sec.~\ref{critical-interaction-aware-trajectory-refinement}.

We use average displacement error (ADE) and final displacement error (FDE) to measure trajectory accuracy, and dynamic time warping (DTW) to evaluate trajectory similarity under temporal variations. We use Jensen--Shannon divergence (JSD) to compare generated and real-world distributions of interaction characteristics. Lower JSD indicates closer agreement.

\hypertarget{main-results}{%
\subsection{Main results}\label{main-results}}

We evaluate the three capabilities introduced in Sec.~\ref{introduction}. The two comparisons with SOTA use the occlusion subset and the full interacting test set at GL.

\begin{table}[!t]
\centering
\caption{Occlusion subset (GL). JSDs and ADE/FDE (m); lower is better.}
\label{tab:occlusion_main}
\begingroup\small
\setlength{\tabcolsep}{2.5pt}
\renewcommand{\arraystretch}{1.12}
\begin{tabular*}{\linewidth}{@{\extracolsep{\fill}}lrrrrrrr@{}}
\toprule
Method & $\mathrm{TTC}_0$ & $\mathrm{RS}_0$ & $T_{\mathrm{adj}}$ & $\mathrm{TTC}_1$ & $\mathrm{RS}_1$ & ADE & FDE \\
\midrule
Ours & \textbf{0.103} & \textbf{0.164} & \textbf{0.317} & \textbf{0.144} & \textbf{0.127} & \textbf{0.780} & \textbf{1.240} \\
SOTA & 0.136 & 0.175 & 0.526 & 0.154 & 0.199 & 0.940 & 1.480 \\
\bottomrule
\end{tabular*}
\endgroup
\end{table}

\begin{figure}[!t]
\centering
\includegraphics[width=\linewidth]{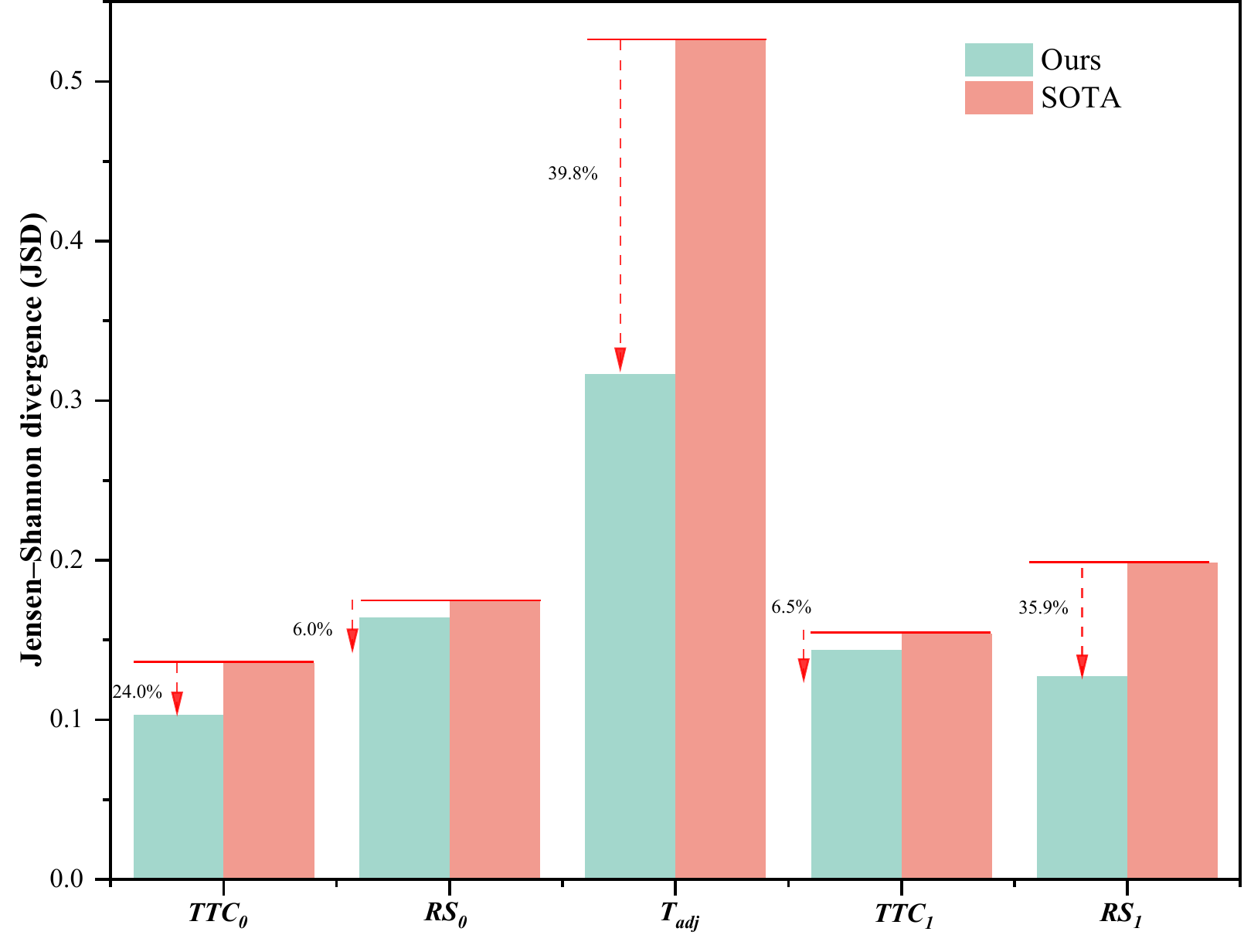}
\caption{Comparison with SOTA on the occlusion subset. Lower JSD indicates closer agreement with real-world data.}
\label{fig:occlusion}
\end{figure}

\textbf{Reproduce occlusion-constrained interaction under partial observability.}
To evaluate this feature, we use an occlusion-specific subset of INTERACTION. It comprises intersection encounters in which the target HDV's view of surrounding traffic is partially obstructed. This evaluation therefore focuses on interaction generation under partial visibility.

Ours achieved lower JSD than SOTA for all five characteristics, together with lower ADE and FDE (Table~\ref{tab:occlusion_main}; Fig.~\ref{fig:occlusion}). These results indicate that closer interaction distributions were achieved alongside accurate vehicle motion. The largest relative gain is in $T_{\mathrm{adj}}$, suggesting that adjustment duration is particularly informative under partial visibility. This distinction matters under occlusion: a path can resemble the recorded geometry while implying a different response to emerging traffic. The FOV ablation examines the contribution of visibility modeling separately.

\begin{figure*}[t]
\centering
\includegraphics[width=\linewidth]{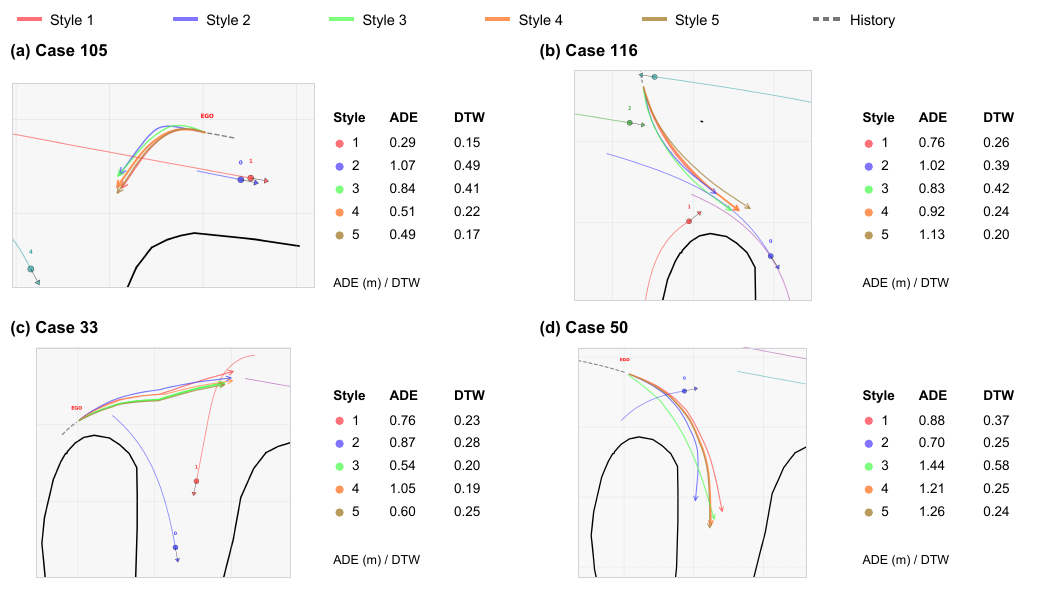}
\caption{Style-conditioned trajectory continuations in four encounters.
Colors indicate selected styles learned from driving data, and dashed lines show the observed trajectory history.}
\label{fig:cases}
\end{figure*}

\textbf{Improve diversity of interactive behaviors.}
The SOTA method generates diverse trajectories by sampling motion tokens without explicitly distinguishing interaction styles. In contrast, our Style-Guided Trajectory Generation module automatically learns distinct interaction styles from real-world driving data through a dynamic expert-activation MoE network. To evaluate this feature, we vary the selected style while keeping the traffic scene and observed history fixed(Fig.~\ref{fig:cases}). 

Different interaction styles lead to distinct driving responses under the same traffic conditions. In Case 105, DTW varies from 0.15 to 0.49 across different styles, reflecting different degrees of similarity to the recorded trajectory. In Case 50, Style 2 follows a smaller turning radius at a lower speed, exhibiting relatively conservative driving behavior. In contrast, Style 1 takes a wider turn at a higher speed, exhibiting relatively aggressive driving behavior. These results demonstrate that our method enables controllable and diverse interaction behaviors within the same driving maneuver.

\textbf{Matching real-world interaction distributions.}
To evaluate this feature, we consider all interaction encounters
with and without occlusion. The Critical-Interaction Residual
Refinement module introduces interaction-specific rewards to
refine generated trajectories toward real-world interaction
distributions.

As shown in Table~\ref{tab:interaction_main} and
Fig.~\ref{fig:sota}, our method reduces JSD by
39.9\%--87.1\% across all five interaction characteristics
compared with SOTA. Notably, JSD for $\mathrm{RS}_1$ and
$\mathrm{TTC}_1$ decreases by 87.1\% and approximately
56\%, respectively. Despite comparable ADE and FDE,
our method more accurately reproduces the interaction
outcomes. These results suggest that our method can
mitigate error accumulation in long-horizon trajectory
generation.

\begin{table}[!t]
\centering
\caption{All interacting GL encounters. JSD and ADE/FDE (m); lower is better.}
\label{tab:interaction_main}
\begingroup\small
\setlength{\tabcolsep}{2.5pt}
\renewcommand{\arraystretch}{1.12}
\begin{tabular*}{\linewidth}{@{\extracolsep{\fill}}lrrrrrrr@{}}
\toprule
Method & $\mathrm{TTC}_0$ & $\mathrm{RS}_0$ & $T_{\mathrm{adj}}$ & $\mathrm{TTC}_1$ & $\mathrm{RS}_1$ & ADE & FDE \\
\midrule
Ours & \textbf{0.013} & \textbf{0.031} & \textbf{0.254} & \textbf{0.025} & \textbf{0.028} & \textbf{0.570} & \textbf{0.980} \\
SOTA & 0.044 & 0.053 & 0.423 & 0.057 & 0.214 & 0.590 & 1.050 \\
\bottomrule
\end{tabular*}
\endgroup
\end{table}

\begin{figure}[!t]
\centering
\includegraphics[width=\linewidth]{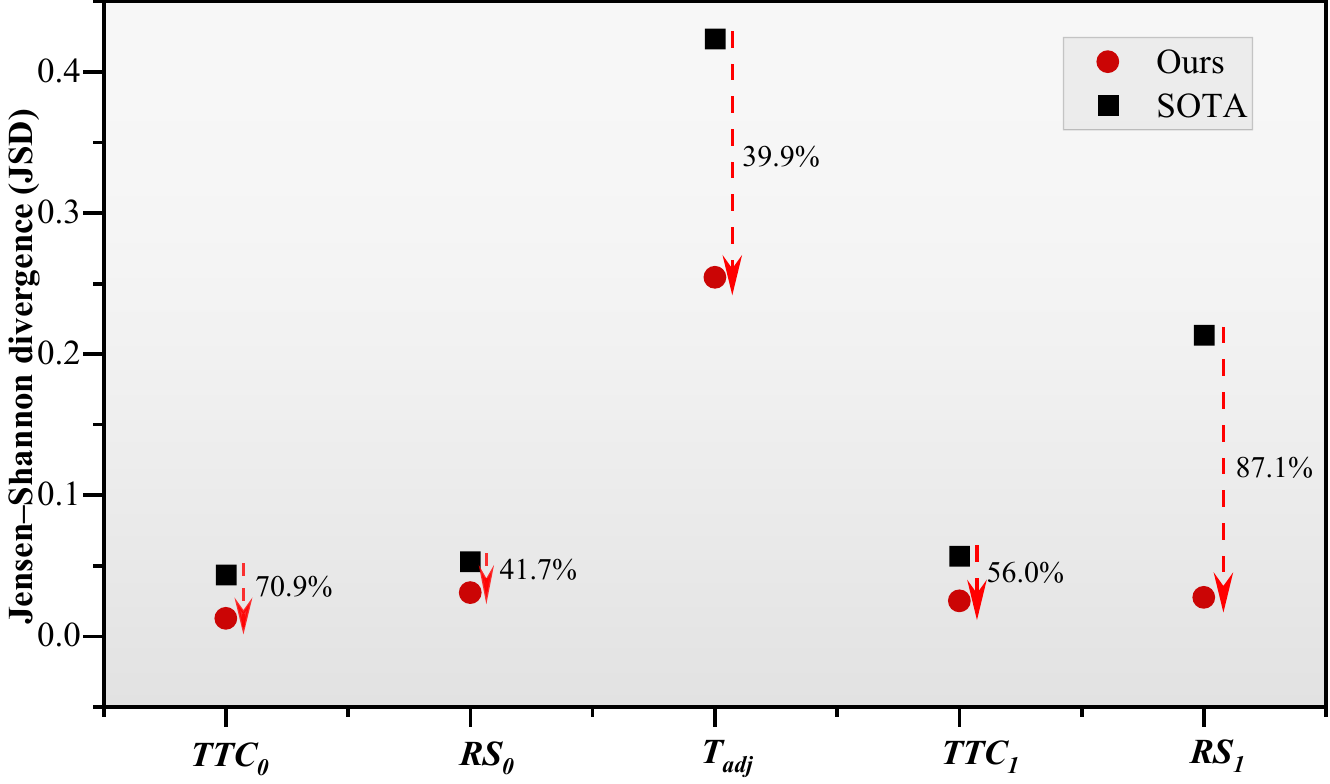}
\caption{Interaction-distribution comparison across all interacting encounters. Lower JSD is better.}
\label{fig:sota}
\end{figure}

\hypertarget{ablation-studies}{%
\subsection{Ablation studies}\label{ablation-studies}}
We conduct ablation studies to assess how each module contributes
to matching real-world interaction distributions. We test the
effect of FOV modeling and style conditioning separately. We
also compare the base generator with two variants of residual
refinement.

\textbf{Occlusion-aware Context Construction.}
As shown in Table~\ref{tab:fov_ablation} and Fig.~\ref{fig:fov},
incorporating FOV reduces JSD for four of the five interaction
characteristics, with reductions of approximately 51\% and 37\%
in $T_{\mathrm{adj}}$ and $\mathrm{RS}_1$, respectively.
The slight increase in $\mathrm{RS}_0$ suggests that FOV modeling
has a limited influence on initial interaction states, with
greater benefits for subsequent driving responses.
These results demonstrate the importance of modeling driver
visibility in reproducing realistic interactions under occlusion.

\begin{table}[!t]
\centering
\caption{FOV ablation on the occlusion subset. All entries are JSD; lower is better.}
\label{tab:fov_ablation}
\begingroup\small
\setlength{\tabcolsep}{2.5pt}
\renewcommand{\arraystretch}{1.12}
\begin{tabular*}{\linewidth}{@{\extracolsep{\fill}}lrrrrr@{}}
\toprule
Method & $\mathrm{TTC}_0$ & $\mathrm{RS}_0$ & $T_{\mathrm{adj}}$ & $\mathrm{TTC}_1$ & $\mathrm{RS}_1$ \\
\midrule
Without FOV & 0.124 & \textbf{0.160} & 0.646 & 0.149 & 0.202 \\
With FOV & \textbf{0.103} & 0.164 & \textbf{0.317} & \textbf{0.144} & \textbf{0.127} \\
\bottomrule
\end{tabular*}
\endgroup
\end{table}

\begin{figure}[!t]
\centering

\includegraphics[width=\linewidth]{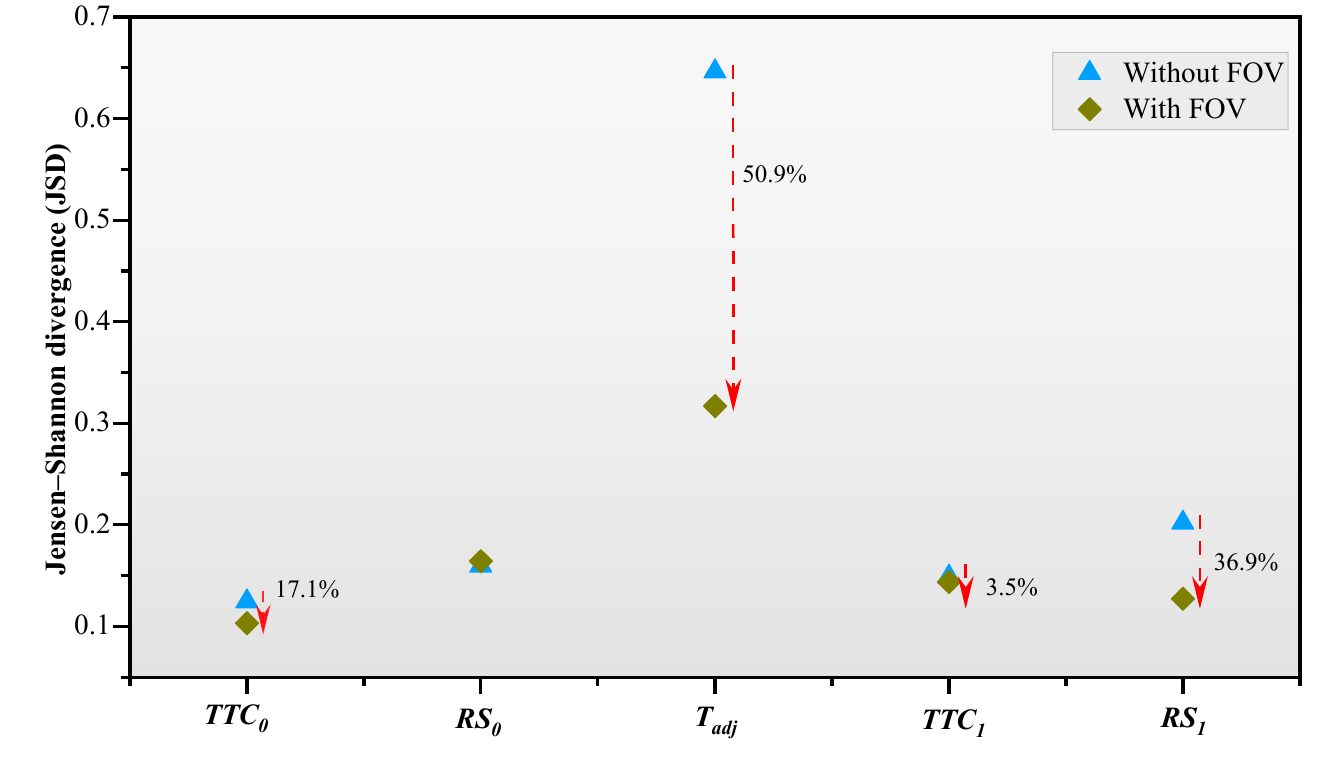}
\caption{Effect of FOV modeling on the occlusion subset.}
\label{fig:fov}
\end{figure}

\textbf{Style-Guided Trajectory Generation.}
As shown in Table~\ref{tab:style_ablation} and
Fig.~\ref{fig:style_ablation}, incorporating the style module
reduces JSD across all five interaction characteristics,
with reductions of over 84\% in $\mathrm{TTC}_0$,
$\mathrm{TTC}_1$, and $\mathrm{RS}_1$.
These results demonstrate that modeling diverse interaction
styles improves the realism of both initial and final
interaction states.

\begin{table}[!t]
\centering
\caption{Interaction-style ablation. All entries are JSD; lower is better.}
\label{tab:style_ablation}
\begingroup\small
\setlength{\tabcolsep}{2.5pt}
\renewcommand{\arraystretch}{1.12}
\begin{tabular*}{\linewidth}{@{\extracolsep{\fill}}lrrrrr@{}}
\toprule
Method & $\mathrm{TTC}_0$ & $\mathrm{RS}_0$ & $T_{\mathrm{adj}}$ & $\mathrm{TTC}_1$ & $\mathrm{RS}_1$ \\
\midrule
With style module & \textbf{0.013} & \textbf{0.031} & \textbf{0.254} & \textbf{0.025} & \textbf{0.028} \\
Without style module & 0.083 & 0.077 & 0.274 & 0.162 & 0.194 \\
\bottomrule
\end{tabular*}
\endgroup
\end{table}

\begin{figure}[!t]
\centering

\includegraphics[width=\linewidth]{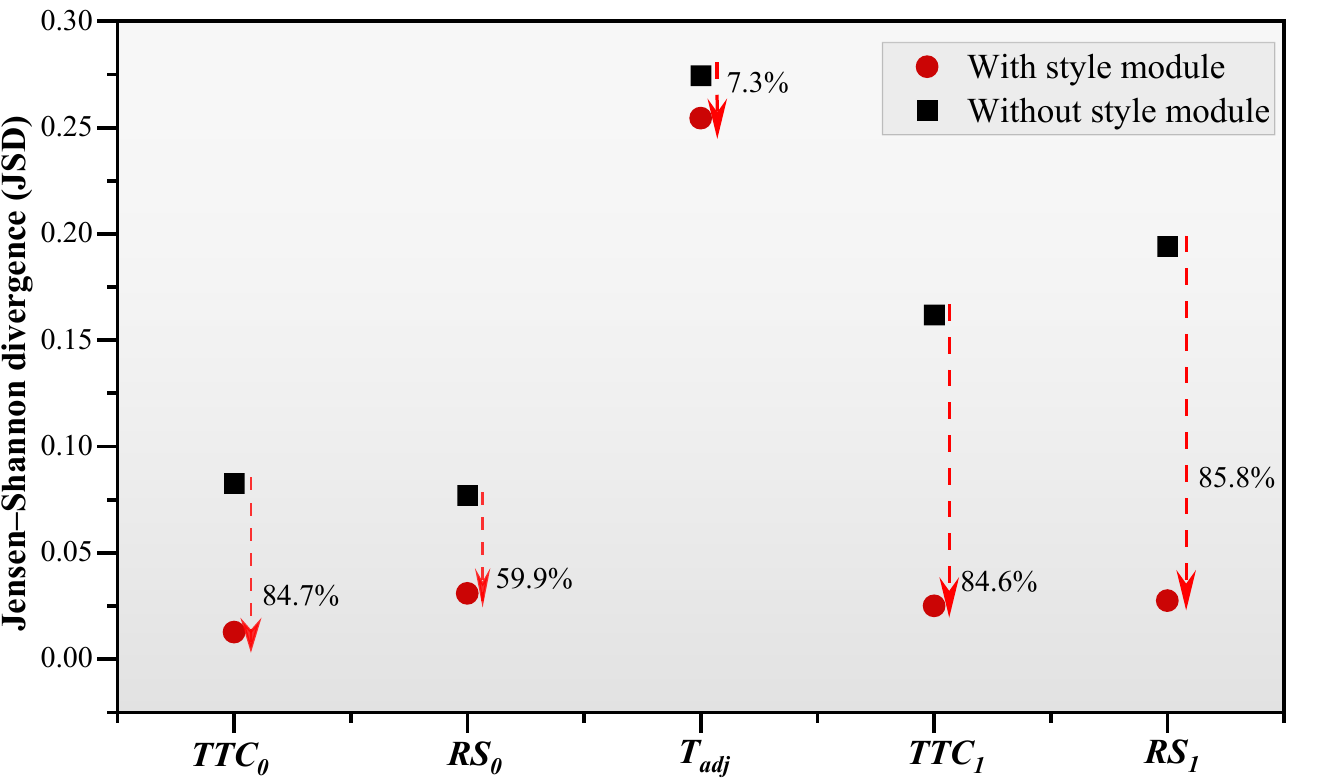}
\caption{Effect of the interaction-style module.}
\label{fig:style_ablation}
\end{figure}

\textbf{Critical-Interaction Residual Refinement.}
As shown in Table~\ref{tab:rl_ablation} and
Fig.~\ref{fig:refinement}, GRPO-based residual refinement
reduces JSD across all five interaction characteristics
compared with imitation learning alone, with the largest
reduction of approximately 96\% in $\mathrm{RS}_1$.
Compared with PPO, GRPO achieves an average JSD reduction
of over 48\% across the five characteristics.
These results demonstrate the effectiveness of residual
refinement in learning diluted interaction signals and
reproducing realistic safety-critical interactions.

\begin{table}[!t]
\centering
\caption{Critical-interaction residual refinement after imitation learning. All entries are JSD; lower is better.}
\label{tab:rl_ablation}
\begingroup\small
\setlength{\tabcolsep}{2.5pt}
\renewcommand{\arraystretch}{1.12}
\begin{tabular*}{\linewidth}{@{\extracolsep{\fill}}lrrrrr@{}}
\toprule
Method & $\mathrm{TTC}_0$ & $\mathrm{RS}_0$ & $T_{\mathrm{adj}}$ & $\mathrm{TTC}_1$ & $\mathrm{RS}_1$ \\
\midrule
IL & 0.121 & 0.174 & 0.717 & 0.133 & 0.657 \\
PPO & 0.027 & \textbf{0.024} & 0.572 & 0.159 & 0.157 \\
GRPO (Ours) & \textbf{0.013} & 0.031 & \textbf{0.254} & \textbf{0.025} & \textbf{0.028} \\
\bottomrule
\end{tabular*}
\endgroup
\end{table}

\begin{figure}[!t]
\centering
\includegraphics[width=\linewidth]{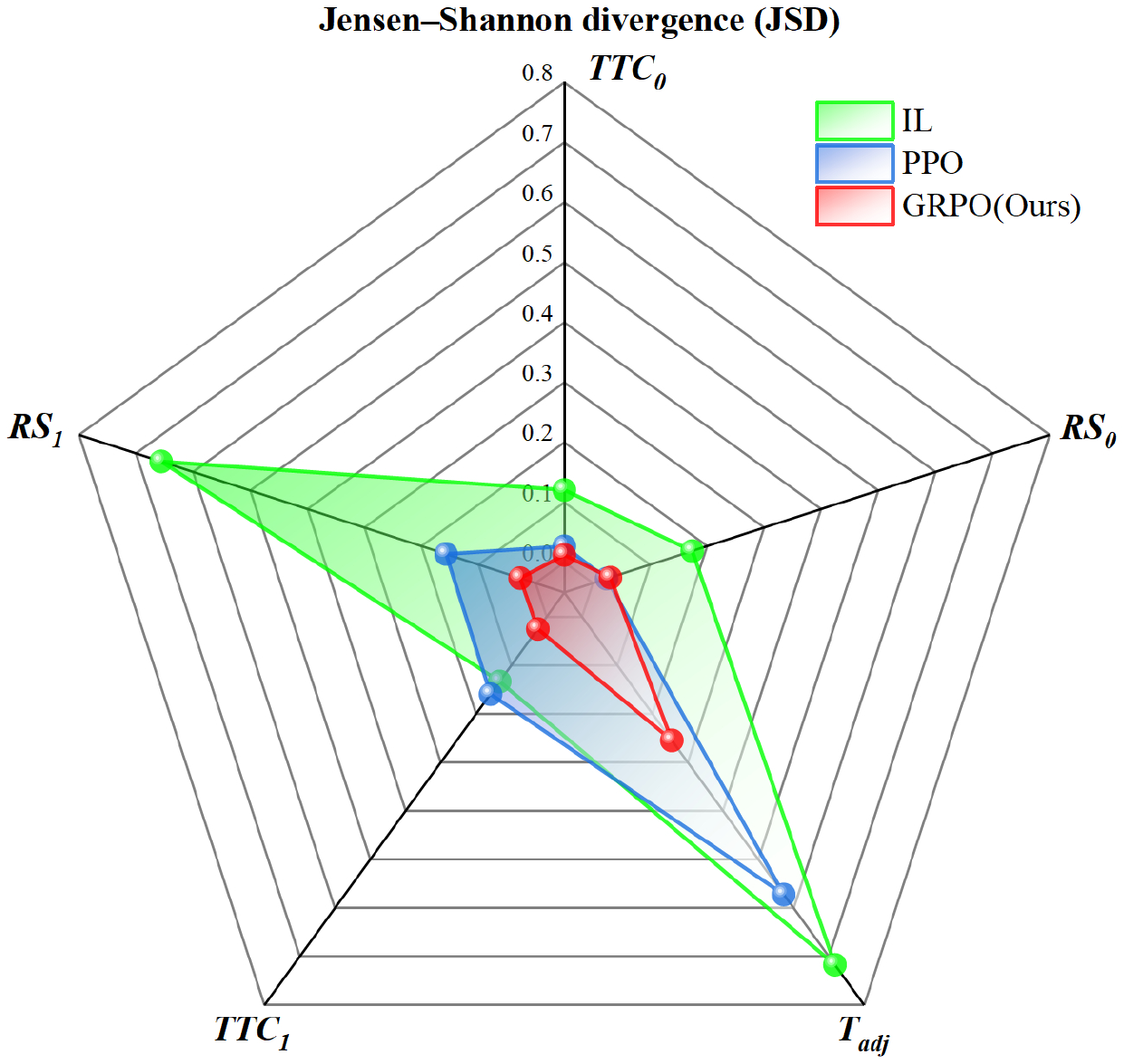}
\caption{IL, PPO, and GRPO (Ours) compared using JSD. Lower is better.}
\label{fig:refinement}
\end{figure}

\hypertarget{generalization}{%
\subsection{Generalization}\label{generalization}}

We evaluate the transferability of learned interaction
behaviors from MA to GL (Fig.~\ref{fig:maps}). All models
are evaluated on GL. SOTA and Ours are trained directly
on GL, while Ours$^{*}$ is trained exclusively on MA.
Ours$^{**}$ is obtained by fine-tuning Ours$^{*}$ on GL
for only one epoch.

As shown in Table~\ref{tab:generalization}, Ours$^{*}$
achieves lower JSD than SOTA across all five interaction
characteristics. After only one epoch of fine-tuning,
Ours$^{**}$ further reduces JSD across all five
characteristics, approaching the performance of full
training on GL. These results demonstrate that our
method generalizes well to unseen intersections and
enables rapid adaptation with minimal fine-tuning.
\begin{figure}[!t]
\centering
\includegraphics[width=\linewidth]{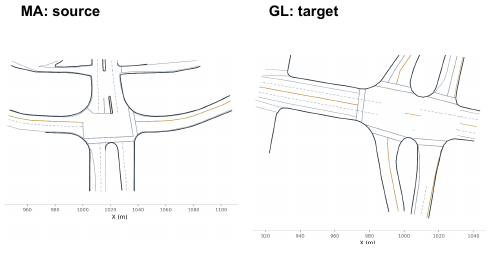}
\caption{Source intersection (MA) and target intersection (GL) for cross-intersection generalization.}
\label{fig:maps}
\end{figure}

\begin{table}[!t]
\centering
\caption{JSD for all configurations evaluated at GL.
SOTA and Ours use GL for training.
Ours$^{*}$ is trained on MA only.
Ours$^{**}$ denotes Ours$^{*}$ fine-tuned on GL for one epoch.}
\label{tab:generalization}
\begingroup\small
\setlength{\tabcolsep}{2.5pt}
\renewcommand{\arraystretch}{1.12}
\begin{tabular*}{\linewidth}{@{\extracolsep{\fill}}lrrrrr@{}}
\toprule
Method & $\mathrm{TTC}_0\downarrow$ & $\mathrm{RS}_0\downarrow$ & $T_{\mathrm{adj}}\downarrow$ & $\mathrm{TTC}_1\downarrow$ & $\mathrm{RS}_1\downarrow$ \\
\midrule
SOTA & 0.044 & 0.053 & 0.423 & 0.057 & 0.214 \\
Ours & 0.013 & 0.031 & 0.254 & 0.025 & 0.028 \\
Ours$^{*}$ & 0.032 & 0.043 & 0.313 & 0.047 & 0.124 \\
Ours$^{**}$ & 0.018 & 0.036 & 0.298 & 0.041 & 0.052 \\
\bottomrule
\end{tabular*}
\endgroup
\end{table}

\par\endgroup

\section{Conclusion}

We propose a generative AI agent to emulate interactive human-driven vehicles for automated driving testing in urban areas. The agent can reproduce occlusion-constrained interaction under partial observability, improve diversity of interactive behaviors, and match real-world interaction distributions. Experiments on the INTERACTION dataset reveal that similar trajectory accuracy does not necessarily imply similar interaction realism. Despite comparable ADE and FDE, our agent reduces JSD by 39.9\%--87.1\%, with particularly pronounced improvements in final-stage interaction characteristics. Different learned styles also produce conservative and aggressive driving responses within the same maneuver. Ablation studies confirm the effectiveness of each proposed component, particularly the importance of residual refinement. Moreover, the agent generalizes to unseen intersections and approaches the performance of full training after only one epoch of fine-tuning. Future work will integrate the proposed HDV agent into a closed-loop driving world model to generate realistic driving videos involving diverse safety-critical interactions. The HDV agent will respond to ego vehicle actions, while the world model will render the evolving traffic states into video sequences for closed-loop automated driving safety testing.

{
    \small
    \bibliographystyle{ieeenat_fullname}
    \bibliography{main}
}

\end{document}